\documentclass[11pt,letterpaper]{article}

\usepackage[letterpaper,margin=1in]{geometry}
\usepackage[T1]{fontenc}
\usepackage[utf8]{inputenc}
\usepackage{lmodern}
\usepackage{microtype}
\usepackage{amsmath}
\usepackage{amssymb}
\usepackage{amsthm}
\usepackage{bm}

\usepackage{graphicx}
\graphicspath{{figures/}}
\usepackage{booktabs}
\usepackage{array}
\usepackage{multirow}
\usepackage{tabularx}
\usepackage{caption}
\usepackage{float}

\usepackage{titlesec}
\titleformat{\section}{\normalfont\Large\bfseries}{\thesection}{0.6em}{}
\titleformat{\subsection}{\normalfont\large\bfseries}{\thesubsection}{0.6em}{}
\titleformat{\subsubsection}{\normalfont\normalsize\bfseries}{\thesubsubsection}{0.6em}{}
\titlespacing*{\section}{0pt}{1.6em}{0.8em}
\titlespacing*{\subsection}{0pt}{1.2em}{0.6em}
\titlespacing*{\subsubsection}{0pt}{0.9em}{0.4em}

\usepackage[numbers,sort&compress]{natbib}
\usepackage{xcolor}
\definecolor{linkblue}{RGB}{28,89,158}
\usepackage[colorlinks=true,linkcolor=black,citecolor=linkblue,urlcolor=linkblue]{hyperref}

\newcommand{\kwta}{\operatorname{k\text{-}WTA}}
\newcommand{\softmax}{\operatorname{softmax}}
\newcommand{\onehot}{\operatorname{onehot}}
\newcommand{\normalizeop}{\operatorname{normalize}}
\newcommand{\retrieveop}{\operatorname{retrieve}}
\newcommand{\bigramop}{\operatorname{bigram}}
\newcommand{\errop}{\operatorname{err}}
\newcommand{\logitsop}{\operatorname{logits}}
\newcommand{\retB}{\operatorname{ret}_B}
\newcommand{\retR}{\operatorname{ret}_R}

\title{
\vspace{-1.5em}
\LARGE\bfseries
The Art of Not Forgetting\\[0.3em]
\Large A Local Learning Architecture for Continual Learning
\vspace{-0.6em}
}

\author{
Ashmith Atmuri$^{*}$ \qquad
Yashaswini Rao Bhogarajula\\[0.7em]
\normalsize\textit{Arkadhi Research}\\[0.7em]
\small
$^{*}$Corresponding author:
\texttt{founder@arkadhi.com}\\[0.25em]
\texttt{yashaswini.rao@arkadhi.com}
}

\date{}

\begin{document}
\maketitle

\begin{abstract}
\noindent
We introduce CMP (Cognitive Memory Primitive), a continual-learning architecture that represents inputs as sparse relational codes, stores them in a two-tier competitive memory, and learns through local updates without end-to-end backpropagation through its feature-generating system. We investigate whether combining sparse representations, local learning, and persistent memory can reduce catastrophic forgetting relative to conventional backpropagation-based continual-learning approaches. On a controlled domain-incremental byte-level language modeling protocol, CMP demonstrates substantially lower backward transfer than a parameter-matched Transformer trained with online Elastic Weight Consolidation (EWC). Across a three-seed replicated 15-domain experiment, CMP exhibits stable forgetting behavior, while separate head-to-head comparisons and domain-order analyses show consistently lower forgetting than the evaluated Transformer baseline under the reported experimental settings. We report these findings alongside a substantial single-domain accuracy gap relative to the Transformer, a null result on a vision benchmark, and a documented failure to combine CMP with an independent accuracy-improving mechanism, reflecting our commitment to reporting both positive and negative outcomes. These results suggest that the combination of sparse representations, local learning, and persistent memory is a promising direction for continual learning, while motivating further investigation into the respective roles of learning rules, representations, and architectural design in mitigating catastrophic forgetting.

\end{abstract}

\vspace{0.3em}

\section{Introduction}
\label{sec: introduction}

Catastrophic forgetting, the tendency of a model trained sequentially on new tasks or domains to overwrite previously acquired knowledge, remains one of the central challenges in continual learning. Existing approaches typically mitigate forgetting through mechanisms such as experience replay, parameter regularization (e.g., Elastic Weight Consolidation (EWC)~\citep{kirkpatrick2017overcoming}), or architectural expansion (e.g., Progressive Networks~\citep{rusu2016progressive}). These methods have demonstrated substantial progress, but they generally operate by augmenting conventional backpropagation-based learning with additional mechanisms that preserve previously learned knowledge.

In this work, we investigate a complementary perspective. Rather than asking how forgetting can be corrected after it emerges, we ask whether an architecture built around local learning, sparse representations, and persistent memory can exhibit greater resistance to catastrophic forgetting without relying on replay buffers or gradient-based importance estimates. Our goal is not to argue that backpropagation is solely responsible for catastrophic forgetting, but to evaluate whether an alternative learning framework can provide improved continual-learning behavior under controlled experimental conditions.

To study this question, we introduce CMP (Cognitive Memory Primitive), a continual-learning architecture that combines established components from several research traditions. CMP integrates sparse relational binding~\citep{plate1995holographic,pollack1990recursive}, competitive content-addressed memory inspired by Sparse Distributed Memory~\citep{kanerva1988sparse} and modern Hopfield networks~\citep{ramsauer2021hopfield}, predictive-coding principles~\citep{rao1999predictive,whittington2017approximation}, and local learning through a weight-local delta-rule readout. None of these individual components is presented as novel. Instead, the contribution of this work lies in their integration into a unified continual-learning system together with a movement-based plasticity regulation mechanism that does not require end-to-end gradient propagation, and in the empirical evaluation of whether this combination reduces catastrophic forgetting relative to a backpropagation-based continual-learning baseline.

\subsubsection*{We state plainly, here, what this paper does not claim.}
It does not claim CMP is more accurate than a Transformer; it is not, by a substantial margin (Section 6). It does not claim general superiority over attention-based architectures. It claims a single, narrower, falsifiable result: on a controlled domain-incremental protocol, with matched architecture size and matched data budget, CMP's backward transfer is substantially better than a Transformer trained with online EWC, and the effect survives a domain-order control and three-seed replication.

If continual-learning performance can be improved through architectural organization and local learning dynamics rather than replay or gradient-derived parameter importance alone, this would suggest an additional direction for designing continual-learning systems. Such an approach could complement existing continual-learning methods while broadening the range of learning mechanisms explored beyond conventional gradient-based optimization.

\subsubsection*{Scope of this work}

We deliberately make limited claims. This paper does not claim that CMP is more accurate than Transformer-based language models; our experiments show a substantial accuracy gap in favor of the Transformer. It does not claim that local learning is universally superior to backpropagation, nor that the individual components of CMP are themselves novel. Instead, the central claim is narrower and directly testable: under the controlled domain-incremental protocol evaluated in this work, CMP demonstrates substantially lower catastrophic forgetting than a parameter-matched Transformer trained with online Elastic Weight Consolidation (EWC). We further report the robustness of this observation through domain-order analyses and replicated experiments, while also documenting negative results, including a null result on Split-MNIST and an unsuccessful attempt to combine CMP with a separate mechanism that improves predictive accuracy.

\subsection{Contributions}

\label{sec:introduction}

The primary contributions of this work are:

\begin{itemize}
\item We introduce CMP (Cognitive Memory Primitive), a continual-learning architecture that combines sparse relational representations, competitive memory, predictive coding, and local learning into a unified framework for continual learning.

\item We present an empirical evaluation on a domain-incremental byte-level language modeling benchmark spanning 15 real-world text domains, demonstrating substantially lower catastrophic forgetting than a parameter-matched Transformer trained with online Elastic Weight Consolidation (EWC) under the reported experimental protocol.

\item We introduce a movement-based plasticity regulation mechanism for local learning that regulates parameter updates without relying on gradient-derived importance estimates, making it compatible with CMP's learning framework.

\item We evaluate the stability of the observed forgetting behavior across multiple random seeds and domain orders, providing a more robust assessment of continual-learning performance.

\item We report both positive and negative findings, including a substantial predictive accuracy gap relative to the Transformer baseline, a null result on Split-MNIST, and an unsuccessful architectural extension, in the interest of transparent and reproducible scientific reporting.

\end{itemize}
\section{Related Work}
\label{sec:related}

\paragraph{Sparse coding.}
Olshausen and Field~\citep{olshausen1996emergence} showed that a sparse coding objective applied to natural images recovers receptive fields resembling primary visual cortex, establishing sparsity as a biologically-grounded representational choice. CMP's k-winners-take-all sparsification throughout (Section 3.1) inherits this motivation directly.

\paragraph{Binding and compositional representation.}
The multiplicative binding operation (Section 3.1) belongs to the Vector Symbolic Architecture family: Pollack's Recursive Distributed Representations~\citep{pollack1990recursive} and Plate's Holographic Reduced Representations~\citep{plate1995holographic} both address representing compositional structure in fixed-width distributed vectors via binding operations. Neither the binding operation nor its use of superposition is new; what is combined with it (Section 3.2--3.7) is specific to this work.

\paragraph{Associative memory.}
Kanerva's Sparse Distributed Memory~\citep{kanerva1988sparse} is the direct ancestor of CMP's competitive, content-addressed memory (Section 3.2). Ramsauer et al.~\citep{ramsauer2021hopfield} establish that modern Hopfield retrieval is mathematically continuous with self-attention; we position CMP's memory as a sibling within that same associative-memory family, distinguished by sparsity and recurrent, one-cue-at-a-time retrieval rather than dense, simultaneous all-pairs lookup, not as a departure from it.

\paragraph{Predictive coding.}
Rao and Ballard~\citep{rao1999predictive} propose predictive coding as a functional account of cortical feedback; Whittington and Bogacz~\citep{whittington2017approximation} show a predictive-coding network can approximate error backpropagation using only local Hebbian updates. CMP's predictive-coding cascade (Section 3.4) uses the mechanism for its stated purpose, local, self-supervised top-down prediction, rather than as an approximation to backpropagation.

\paragraph{Complementary learning systems.}
McClelland, McNaughton, and O'Reilly~\citep{mcclelland1995why} argue the hippocampus and neocortex form complementary systems precisely to avoid the interference that a single fast-learning system would suffer: the direct theoretical motivation for pairing a sparse, fast-writing memory with a slower, distributed readout in CMP, and for expecting that pairing to resist forgetting for structural reasons.

\paragraph{Continual learning.}
EWC~\citep{kirkpatrick2017overcoming} regularises parameter movement by Fisher-information-weighted importance; Synaptic Intelligence~\citep{zenke2017continual} computes importance online from the training trajectory; Progress \& Compress~\citep{schwarz2018progress} introduces the online-EWC variant used as our baseline (Section 4.3). Progressive Networks~\citep{rusu2016progressive}, instead, avoid architectural forgetting by freezing old columns and adding new ones per task. iCaRL~\citep{rebuffi2017icarl} represents the replay-based family, noted but not compared directly. Lopez-Paz and Ranzato~\citep{lopezpaz2017gradient} define the BWT/FWT metrics used throughout Section 4, adapted here to a bits-per-byte formulation with an explicitly stated sign convention.

\paragraph{Non-attention sequence architectures.}
Mamba~\citep{gu2023mamba} and related state-space models achieve strong long-context efficiency without attention, but, to our knowledge, and consistent with our own accuracy results (Section 6), no non-attention architecture has cleanly matched Transformer accuracy at matched parameters and compute on short-context tasks. We note this as precedent for reporting a real accuracy gap alongside a different, real contribution.

\paragraph{Byte-level language modelling.}
The Byte Latent Transformer~\citep{pagnoni2025byte} allocates compute dynamically based on next-byte entropy, operating, like CMP, directly on bytes rather than a learned subword vocabulary. We note the shared byte-level premise as context for Section 4's bits-per-byte metric.

\section{Architecture}
\label{sec:architecture}

Before presenting the mathematical formulation, we provide a high-level overview of the architecture. A pair of input bytes is first transformed into a sparse relational representation (Section~3.1), which is stored in a two-tier competitive memory for subsequent retrieval (Section~3.2). A higher-level representation is then constructed and used to predict the lower-level representation through a predictive-coding hierarchy (Sections~3.3--3.4). The resulting representations, memory retrievals, and prediction errors are combined by a readout module to produce the next-byte prediction (Section~3.5). The readout is trained using a local learning rule without end-to-end backpropagation through the preceding stages (Section~3.6). Finally, a movement-based plasticity regulation mechanism limits parameter updates according to local weight dynamics (Section~3.7). Section~3.8 compares the resulting design with self-attention along the architectural dimensions motivating CMP.

Figure 1 provides a high-level view of the complete CMP pipeline. The following subsections explain each component in order.

\begin{figure}[H]
  \centering
  \includegraphics[width=0.55\linewidth]{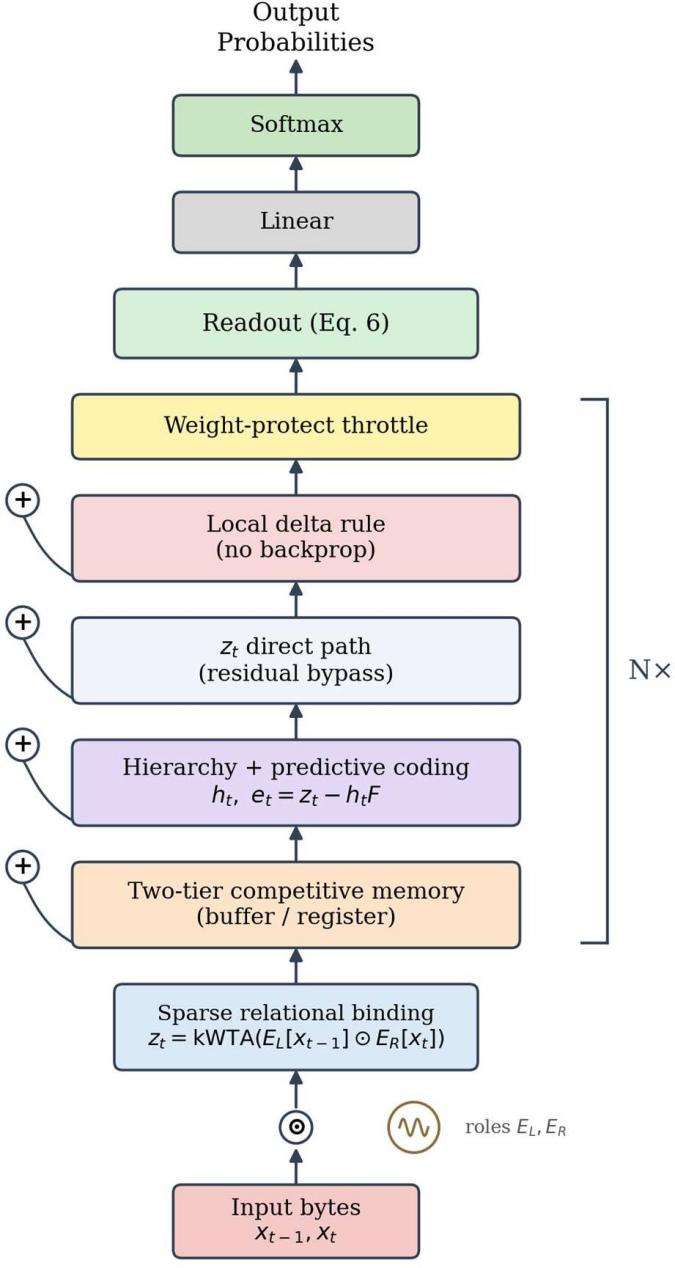}
  \caption{Overall architecture of CMP. Input bytes are transformed into sparse relational representations, stored in a competitive memory, processed through a predictive-coding hierarchy, and decoded using a locally trained readout with movement-based plasticity regulation.}
  \label{fig:cmp-pipeline}
\end{figure}

We describe the architecture in the same order Vaswani et al.~\citep{vaswani2017attention} describe self-attention: the core binding operation first, then how it composes into memory, hierarchy, and a learning rule. A small version of this figure, with the relevant piece highlighted, appears at the start of each subsection below so to aid readability; each subsection begins with a reduced version of Figure~\ref{fig:cmp-pipeline}, highlighting the component under discussion.

\subsection{Sparse Relational Representation}
\label{subsec:binding}

\begin{center}
  \includegraphics[width=0.42\linewidth]{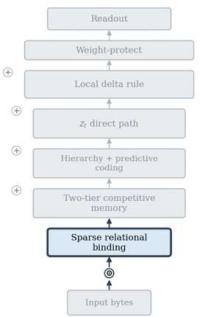}
\end{center}

The first stage of CMP transforms each pair of consecutive input bytes into a fixed-dimensional relational representation that serves as the basic unit processed throughout the remainder of the architecture. Rather than operating directly on raw byte values, CMP constructs a sparse representation that can be stored, retrieved, and composed by later modules while remaining independent of vocabulary size. Given the previous byte $x_{t-1}$ and the current byte $x_t$, this stage produces a sparse relational vector $z_t$.

Two fixed embedding matrices, $E_L$ and $E_R$, assign each byte to left-role and right-role representations, respectively. These role-specific embeddings are combined through elementwise multiplication, followed by k-winners-take-all sparsification and $\ell_2$ normalization:

\begin{equation}
\label{eq:binding}
z_t \;=\; \normalizeop\!\bigl(\kwta_k\!\bigl(E_L[x_{t-1}] \;\odot\; E_R[x_t]\bigr)\bigr)
\end{equation}

where $\odot$ denotes elementwise multiplication, $\kwta_k(\cdot)$ retains the $k$ largest activations while suppressing all remaining entries, and $\normalizeop(\cdot)$ performs $\ell_2$ normalization. The resulting vector $z_t \in \mathbb{R}^{r}$ is the sparse relational representation associated with position $t$ and serves as the fundamental representation throughout the CMP architecture.

The embedding matrices $E_L$ and $E_R$ are randomly initialized once and remain fixed throughout training. Consequently, learning occurs only within the downstream components of the architecture, while the input representation remains stable across the learning process. This design reduces the number of trainable parameters and isolates the effects of the subsequent memory and learning mechanisms. The implications of using fixed embeddings are discussed further as a limitation in Section~6.

All subsequent components operate on the relational representation $z_t$. The next subsection describes how these representations are stored and retrieved using CMP's competitive memory, after which they are transformed through the hierarchical predictive-coding mechanism before contributing to the final prediction.

\subsection{Two-tier competitive memory}
\label{subsec:memory}

\begin{center}
    \includegraphics[width=0.42\linewidth]{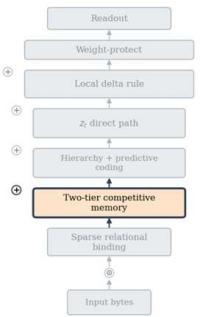}
\end{center}

The sparse relational representation $z_t$ captures local context between consecutive input bytes but, by itself, provides no mechanism for retaining information beyond the current input pair. CMP therefore augments relational binding with a two-tier competitive memory that stores previously observed representations and retrieves them through content-based addressing. This memory enables information from earlier inputs to influence future predictions while maintaining a fixed-size external memory.

The memory consists of two components: a fast buffer ($B$), which captures recently observed representations, and a slower register ($R$), which stores representations that remain useful over longer time scales. Given the current relational representation $z_t$, retrieval is performed as a similarity-weighted combination of the stored memory vectors:

\begin{equation}
\label{eq:retrieval}
\retrieveop(M, z_t)
=
\sum_{i=1}^{|M|}
\softmax_i\!\left(
\tau\,\hat{z}_t \cdot \hat{m}_i
\right)
m_i
\end{equation}

where $M$ denotes the queried memory (buffer, register, or their combination), $\hat{z}_t$ and $\hat{m}_i$ are $\ell_2$-normalized vectors, and $\tau$ controls the sharpness of the similarity weighting. The retrieved vector is therefore a weighted combination of the memory slots whose contents most closely match the current relational representation.

Memory writes are competitive rather than unconditional. A memory slot is updated only when its similarity to the incoming representation exceeds a calibrated matching threshold, reducing interference between unrelated representations and preventing updates caused by weak or coincidental overlap. The matching threshold is defined as

\begin{equation}
\label{eq:write}
\theta_{\mathrm{match}}(r,k,s)
=
0.86\,
\mathbb{E}
\left[
\max_{i\neq j}
\,
\hat{v}_i \cdot \hat{v}_j
\right].
\end{equation}

The expectation is estimated over randomly generated sparse vectors having the same dimensionality and sparsity as the representations used by CMP. The buffer size, register size, and sparsity level are derived from the representation dimension $r$ according to the fixed scaling rules described in Appendix~A rather than being tuned independently. We discuss the implications of this design choice as a limitation in Section~6.

\begin{figure}[t]
    \centering
    \includegraphics[width=0.70\linewidth]{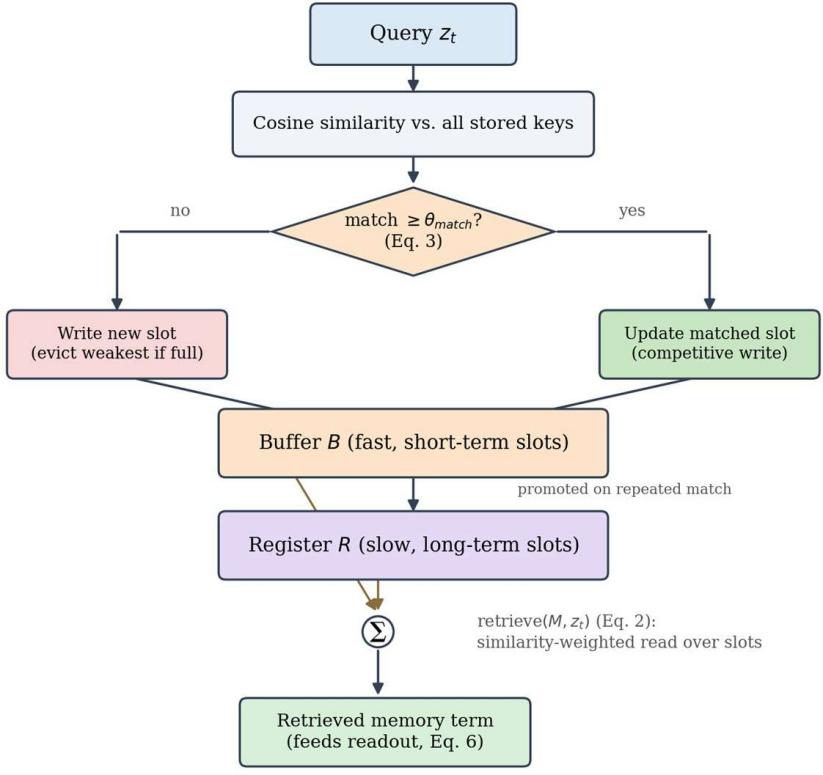}
    \caption{Overview of the two-tier competitive memory. Incoming relational representations are first matched against existing memory slots. Successful matches update existing slots, while persistent representations are promoted from the fast buffer to the long-term register. Retrieval is performed through similarity-weighted aggregation over stored representations.}
    \label{fig:memory}
\end{figure}

Figure~\ref{fig:memory} summarizes the complete memory pipeline, including competitive write decisions, promotion from the fast buffer to the long-term register, and similarity-based retrieval.

Separating the buffer and register allows CMP to distinguish short-term adaptation from longer-term storage. Representations that repeatedly satisfy the matching criterion are promoted from the fast buffer into the register, whereas transient representations remain confined to short-term memory. The retrieved memory representation is then passed to the hierarchical prediction stage described in the following subsection.
\subsection{Predictive coding}
\label{subsec:predictive}

\begin{center}
    \includegraphics[width=0.42\linewidth]{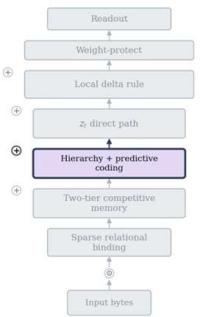}
\end{center}

The hierarchical representation introduced in the previous subsection provides a coarse summary of the current input, but it does not indicate which aspects of the underlying relational representation remain unexplained. CMP therefore incorporates a predictive-coding stage that models the relationship between coarse and fine representations. Rather than using the higher-level representation directly, this stage computes a sparse prediction residual that captures information not explained by the coarse representation.

Given the higher-level representation $h_t$ and the relational representation $z_t$, a locally trained prediction matrix $F$ produces an estimate of the fine representation. The prediction residual is then sparsified and normalized to obtain an additional contextual representation:

\begin{equation}
\label{eq:predictive}
\hat{z}_t
=
h_tF,
\qquad
e_t
=
\normalizeop\!\left(
\kwta_k
\left(
z_t-\hat{z}_t
\right)
\right)
\end{equation}

where $\hat{z}_t$ denotes the predicted relational representation, $F$ is the locally trained prediction matrix, and $e_t$ is the resulting sparse prediction residual. The residual emphasizes components of the relational representation that are not captured by the corresponding higher-level representation, thereby providing complementary information for subsequent processing.

Unlike the final prediction module, the predictive-coding layer is trained exclusively through a local reconstruction objective. Consequently, the parameters of $F$ are updated independently of the downstream prediction loss, allowing this stage to learn predictive relationships without requiring end-to-end backpropagation through the architecture.

At this point, four complementary representations are available: the relational representation $z_t$, the retrieved memory representation, the higher-level representation $h_t$, and the prediction residual $e_t$. The following subsection describes how these signals are integrated to generate the final next-byte prediction.

\subsection{Readout}
\label{subsec:readout}

\begin{center}
    \includegraphics[width=0.42\linewidth]{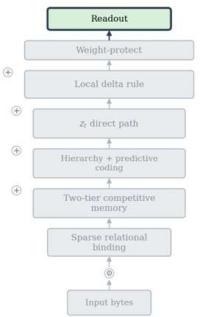}
\end{center}

The preceding stages produce multiple complementary representations, each capturing a different aspect of the input sequence. The objective of the readout stage is to integrate these representations into a unified prediction over the next input byte. Rather than employing a deep nonlinear decoder, CMP uses a linear readout that combines relational, memory, hierarchical, predictive, and temporal-context representations into a single set of output logits.

Specifically, the relational representation $z_t$, the retrieved buffer and register memories ($\retB$ and $\retR$), the higher-level representation $h_t$, the predictive-coding residual $e_t$, and a collection of leaky-integrator context representations (described in Appendix~A) are combined together with a learned bigram table $W_h$ to produce the logits over the 256 possible byte values:

\begin{equation}
\label{eq:readout}
\logitsop_t
=
W_h^{\top}[x_t]
+
b
+
z_tW_0^{\top}
+
\retB W_B^{\top}
+
\retR W_R^{\top}
+
h_tW_{h_2}^{\top}
+
e_tW_{pc}^{\top}
+
\sum_{\alpha}
c_t^{(\alpha)}
W_{s_\alpha}^{\top}
\end{equation}

where $b$ is the output bias term and each weight matrix projects a different representation into the common output space. The resulting logits are converted to a probability distribution over the next byte using the standard softmax function.

The readout itself consists solely of linear projections and summation. All learning therefore occurs through the associated weight matrices rather than through additional hidden layers, allowing the prediction stage to remain compatible with the local learning framework employed throughout CMP.

The following subsection describes how these readout weights are updated using a local learning rule without end-to-end backpropagation through the preceding stages.

\subsection{Local learning rule}
\label{subsec:local-rule}

\begin{center}
    \includegraphics[width=0.42\linewidth]{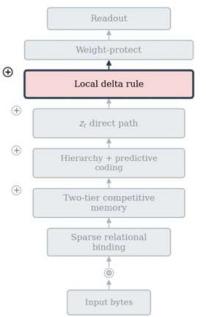}
\end{center}

The readout stage produces the next-byte prediction, but the preceding architecture does not employ end-to-end backpropagation to propagate prediction errors through earlier processing stages. Instead, each readout weight is updated using information available locally at the output layer. Consequently, every readout projection is optimized independently from the representations that generate its input, allowing learning to proceed through a collection of local weight updates.

Let the prediction error be defined as the difference between the target distribution and the predicted probability distribution:

\begin{align}
\label{eq:delta}
\errop_t
&=
\onehot(y_t)
-
\softmax(\logitsop_t)
\\[0.4em]
\label{eq:update}
\Delta W_{\bullet}
&=
\frac{\eta}{N}
\sum_t
\errop_t^{\top}
\phi_{\bullet}(t)
\end{align}

where $y_t$ is the target byte, $\logitsop_t$ denotes the output logits from Eq.~\ref{eq:readout}, $\eta$ is the learning rate, $N$ is the normalization factor corresponding to the minibatch size, and $\phi_{\bullet}(t)$ represents the input representation associated with a particular readout weight matrix. Depending on the projection being updated, $\phi_{\bullet}(t)$ may correspond to the relational representation, retrieved memory, hierarchical representation, predictive-coding residual, or one of the temporal-context representations.

Each readout projection is therefore updated using only its local input activity together with the prediction error computed at the output layer. No gradients are propagated through the relational binding, competitive memory, hierarchical, or predictive-coding modules, and each upstream component is optimized through its own local learning mechanism.

Although this update rule enables local learning, applying identical updates across all parameters throughout continual training can still produce interference between previously acquired and newly learned knowledge. The following subsection introduces the plasticity-regulation mechanism used to mitigate this effect by adapting the magnitude of weight updates during learning.

\subsection{Plasticity regulation}
\label{subsec:weight-protect}

\begin{center}
    \includegraphics[width=0.42\linewidth]{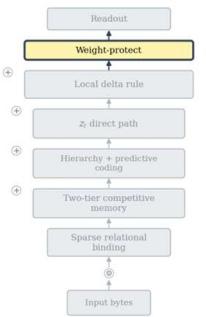}
\end{center}

The local learning rule described in the previous subsection updates each readout projection independently using the current prediction error. While this enables continual adaptation, repeated training across multiple domains can cause parameters that were important for previously learned tasks to be modified by subsequent updates. To reduce such interference, CMP incorporates a plasticity-regulation mechanism that estimates the relative importance of each readout tensor from its accumulated parameter movement and uses this estimate to modulate future updates.

Following the completion of training on domain $i$, the mean absolute parameter movement of each readout tensor is accumulated into an importance estimate:

\begin{align}
\label{eq:importance}
I_{\bullet}
&\leftarrow
I_{\bullet}
+
\mathbb{E}
\left[
\left|
W_{\bullet}^{(i)}
-
W_{\bullet}^{(i-1)}
\right|
\right]
\\[0.4em]
\label{eq:throttle}
\Delta W_{\bullet}
&\leftarrow
\frac{1}
{1+\lambda I_{\bullet}}
\,
\Delta W_{\bullet}
\end{align}

where $I_{\bullet}$ denotes the accumulated importance associated with a particular readout tensor, $W_{\bullet}^{(i)}$ and $W_{\bullet}^{(i-1)}$ are the parameter values after training on consecutive domains, and $\lambda$ controls the strength of the plasticity regulation. Tensors that have exhibited larger cumulative movement receive progressively smaller updates during subsequent training, while tensors with lower accumulated importance remain comparatively more adaptable.

Unlike Elastic Weight Consolidation (EWC), which estimates parameter importance using the Fisher Information Matrix derived from gradients of the log-likelihood, the proposed mechanism relies solely on accumulated parameter movement. This choice is motivated by the local-learning framework employed throughout CMP, where end-to-end gradients are not propagated through the architecture. Consequently, parameter movement provides a gradient-independent estimate of importance that is compatible with the learning dynamics of the model.

The individual components of CMP—including sparse relational binding, competitive memory, hierarchical representations, predictive coding, and the local delta rule—are adapted from established ideas described in Section~2. The primary methodological contribution of this work is the integration of the movement-based importance estimate (Eq.~\ref{eq:importance}) with adaptive update scaling (Eq.~\ref{eq:throttle}) to regulate plasticity during continual learning within a fully local-learning framework.

The following section describes the experimental protocol used to evaluate the proposed architecture and the effectiveness of the plasticity-regulation mechanism under continual language-modeling benchmarks.
\subsection{Comparison with self-attention}
\label{subsec:vs-attention}

The preceding subsections describe the individual components of CMP and how they interact to perform continual sequence modeling using local learning. An important architectural question is why the core representation is constructed through sparse relational binding rather than the self-attention mechanism used in Transformer-based language models. This section summarizes the principal design differences between the two approaches.

\begin{table}[t]
    \centering
    \caption{Qualitative comparison between self-attention and sparse relational binding. The comparison follows the high-level style of the complexity analysis presented by Vaswani et al.~\citep{vaswani2017attention}.}
    \label{tab:vs-attention}
    \small
    \begin{tabular}{@{}lll@{}}
        \toprule
        \textbf{Property} &
        \textbf{Self-attention} &
        \textbf{Sparse relational binding} \\
        \midrule
        Computation per step &
        Pairwise interactions over the context &
        Fixed-size local operation \\
        Learning mechanism &
        End-to-end backpropagation &
        Local learning \\
        Memory &
        Context recomputed at each forward pass &
        Explicit sparse persistent memory \\
        \bottomrule
    \end{tabular}
\end{table}

The distinction between the two approaches is primarily architectural rather than representational. Self-attention constructs representations by simultaneously computing interactions among all tokens within the current context window, whereas sparse relational binding incrementally updates representations through local recurrent operations combined with explicit associative memory. Consequently, CMP emphasizes fixed-size local computation and persistent memory instead of repeated dense interactions across the full input sequence.

Modern Hopfield networks have been shown to possess a close mathematical relationship with Transformer attention mechanisms~\citep{ramsauer2021hopfield}. Accordingly, the associative-memory component of CMP should not be viewed as an alternative computational paradigm, but rather as a related memory-retrieval mechanism operating under different architectural constraints. The principal differences are the use of sparse representations, recurrent one-cue-at-a-time retrieval, explicit long-term memory, and compatibility with locally computed learning updates.

The comparison presented in Table~\ref{tab:vs-attention} is intended to highlight these architectural differences rather than establish superiority of one approach over the other. The experimental evaluation in the following sections assesses whether the proposed design provides advantages for continual language modeling under the evaluation protocol considered in this work.
\section{Continual Learning Protocol and Results}
\label{sec:experiments}

This section evaluates CMP under a domain-incremental continual-learning protocol designed to assess both predictive performance and resistance to catastrophic forgetting. The evaluation proceeds in four stages. First, we verify that the model achieves meaningful language-modeling performance before interpreting forgetting metrics. Second, we examine the reproducibility of the proposed plasticity-regulation mechanism across multiple random seeds. Third, we compare the proposed method with established continual-learning baselines under matched training conditions. Finally, we evaluate robustness to domain ordering and investigate whether the observed behavior extends beyond text to an image-classification setting.

\subsection{Continual-learning protocol}
\label{subsec:protocol}

We adopt a domain-incremental continual-learning protocol in which a single model is trained sequentially across $T$ domains without replaying previously observed data or providing explicit task identifiers. After completing training on each domain, the model is evaluated on both the current domain and all previously encountered domains, producing a $T\times T$ matrix of Bits-Per-Byte (BPB) scores that summarizes continual-learning performance.

This evaluation protocol intentionally removes two common sources of auxiliary information frequently used in continual-learning systems: replay buffers and task identifiers. Consequently, knowledge retention must emerge from the architecture and learning dynamics themselves rather than from revisiting previous training data or conditioning on the active task identity.

To quantify continual-learning behavior, we report Backward Transfer (BWT) and Forward Transfer (FWT), adapted from Lopez-Paz and Ranzato~\citep{lopezpaz2017gradient} for the Bits-Per-Byte metric. Because lower BPB indicates better performance, the sign convention is inverted relative to the original accuracy-based formulation.

\begin{figure}[t]
    \centering
    \includegraphics[width=0.75\linewidth]{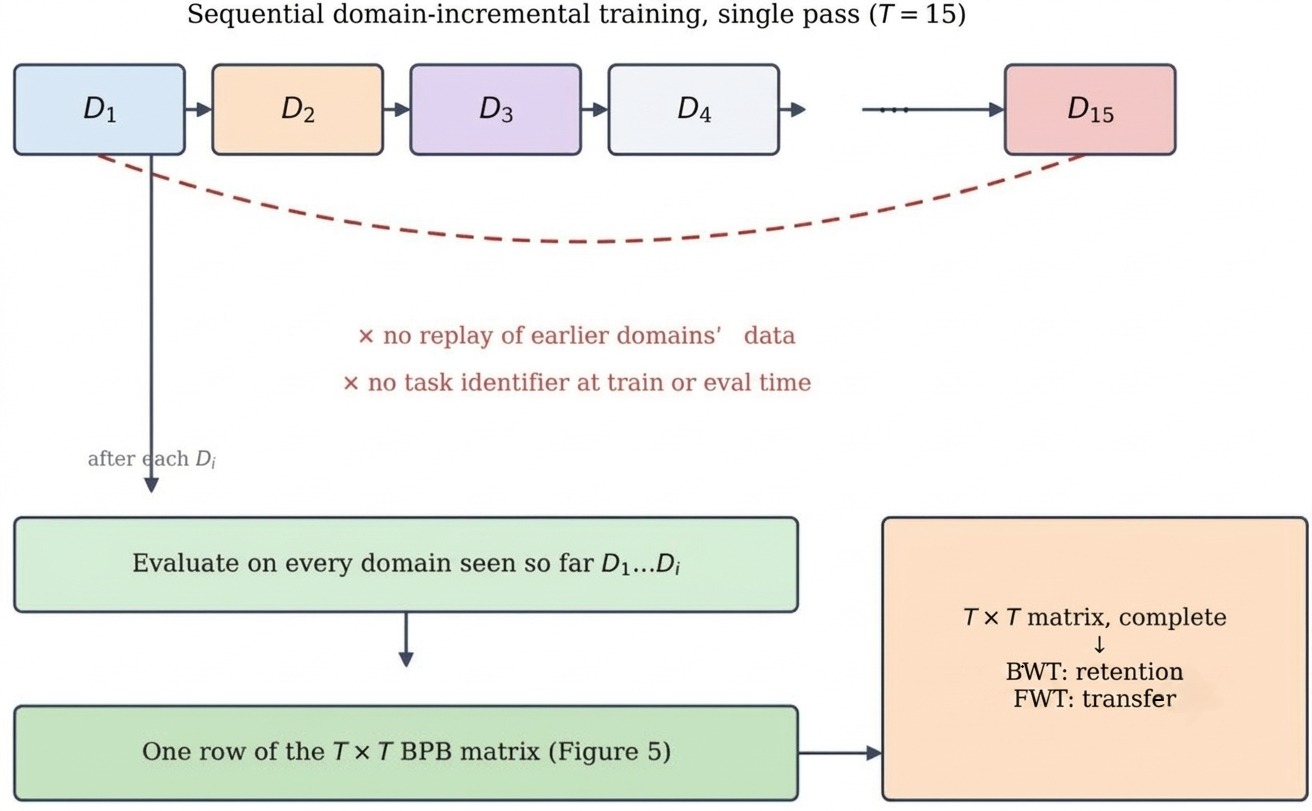}
    \caption{Domain-incremental continual-learning protocol. A single model is trained sequentially across domains without replay or task identifiers. After each training stage, evaluation is performed on all domains encountered up to that point, producing a complete BPB evaluation matrix.}
    \label{fig:protocol}
\end{figure}

\begin{align}
\label{eq:bwt}
\mathrm{BWT}
&=
\frac{1}{T-1}
\sum_{i=1}^{T-1}
\left(
\mathrm{BPB}(i,T)
-
\mathrm{BPB}(i,i)
\right)
\\[0.4em]
\label{eq:fwt}
\mathrm{FWT}
&=
\frac{1}{T-1}
\sum_{i=2}^{T}
\left(
\mathrm{BPB}_{\mathrm{0shot}}(i)
-
\bigramop(i)
\right)
\end{align}

Under this convention, positive BWT indicates forgetting, corresponding to an increase in BPB after subsequent domains have been learned, whereas negative BWT indicates backward transfer. Likewise, negative FWT indicates performance better than the bigram baseline on previously unseen domains.

The evaluation consists of 15 sequential domains spanning encyclopedic text, software source code, literary works, news articles, technical documentation, and multilingual data. Specifically, the corpus includes Wikipedia, CPython source code, Shakespeare, the King James Bible, Reuters news, \textit{Moby Dick}, \textit{The Federalist Papers}, Darwin, Whitman, Lodash source code, \textit{Meditations} by Marcus Aurelius, \textit{Flatland}, \textit{Alice's Adventures in Wonderland}, Python technical documentation, and a Hindi corpus. This collection was assembled specifically for this study and does not constitute a standardized continual-learning benchmark. We therefore use established continual-learning metrics on a custom evaluation corpus and discuss the resulting limitations in Section~\ref{sec:discussion}

\subsection{Accuracy audit}
\label{subsec:audit}

Before interpreting continual-learning metrics, it is necessary to establish that the model has learned meaningful predictive structure. Forgetting measurements are only informative if the model first acquires knowledge that can subsequently be retained or forgotten.

To verify this prerequisite, we compare the post-training performance of CMP against a simple bigram frequency baseline across all evaluation domains. For every domain and across all three random seeds (42, 43, and 44), CMP consistently achieves lower BPB than the corresponding bigram baseline, with improvements ranging from 0.41 to 1.22 BPB. The same verification procedure is applied to all continual-learning baselines included in the subsequent comparisons.

Having established that the model consistently learns beyond the trivial baseline, the following subsection evaluates whether the proposed plasticity-regulation mechanism produces reproducible continual-learning improvements across multiple independent training runs.

\subsection{Main result: three-seed replication}
\label{subsec:replication}

The proposed plasticity-regulation mechanism is intended to improve continual-learning performance independently of random initialization. To evaluate the robustness of the observed effect, we repeat the complete 15-domain continual-learning experiment using three independent random seeds (42, 43, and 44) while keeping the training protocol and hyperparameters unchanged.

\begin{table}[t]
    \centering
    \caption{Continual-learning performance across three independent random seeds using the proposed plasticity-regulation mechanism on the 15-domain evaluation sequence.}
    \label{tab:replication}
    \small
    \begin{tabular}{@{}lcc@{}}
        \toprule
        \textbf{Seed} &
        \textbf{BWT} &
        \textbf{FWT} \\
        \midrule
        42 & $+0.2353$ & $+0.5956$ \\
        43 & $+0.2352$ & $+0.2978$ \\
        44 & $+0.2512$ & $+0.3175$ \\
        \midrule
        Mean $\pm$ Std. &
        $+0.2406 \pm 0.0092$ &
        $+0.4036 \pm 0.1665$ \\
        \bottomrule
    \end{tabular}
\end{table}

Table~\ref{tab:replication} shows that Backward Transfer (BWT) remains highly consistent across all three runs, exhibiting a standard deviation of 0.0092, corresponding to less than 4\% of the mean value. This indicates that the continual-learning behavior associated with the proposed plasticity-regulation mechanism is reproducible under different random initializations.

In contrast, Forward Transfer (FWT) exhibits substantially greater variability across seeds. The higher variance is primarily attributable to seed 42, which achieves a noticeably larger FWT value than the remaining two runs. Consequently, while the observed BWT behavior appears stable across random seeds, the corresponding FWT estimates should be interpreted more cautiously.

These experiments establish that the principal result reported in this work—the reduction in forgetting measured by BWT—is reproducible rather than arising from a single favorable initialization. However, reproducibility alone does not establish competitive performance. The following subsection therefore compares the proposed approach with established continual-learning baselines under matched training conditions.

\subsection{Baseline comparison: naive fine-tuning and online EWC}
\label{subsec:baseline}

To assess whether the observed continual-learning behavior extends beyond comparisons with an unregularized model, we compare CMP against a parameter-matched ($\sim$6.5M parameters) byte-level Transformer under the same domain-incremental evaluation protocol. Two baseline configurations are considered. The first is naive sequential fine-tuning without any continual-learning mechanism. The second is online Elastic Weight Consolidation (Online EWC)~\citep{schwarz2018progress}, which maintains a running approximation of parameter importance and was selected in place of the original EWC formulation~\citep{kirkpatrick2017overcoming} because it scales more naturally to long sequences of domains without requiring storage of separate Fisher Information Matrices for every task.

All models are trained using the identical sequence of domains, optimization schedule, and evaluation procedure described in Section~\ref{subsec:protocol}. Matching the parameter count and training protocol allows differences in continual-learning performance to be attributed primarily to the learning mechanisms rather than to model capacity or computational budget.

The comparison therefore evaluates three conditions:

\begin{itemize}
    \item \textbf{Naive fine-tuning}: sequential training without any mechanism for mitigating catastrophic forgetting.
    \item \textbf{Online EWC}: continual learning using a running Fisher-based estimate of parameter importance following Schwarz et al.~\citep{schwarz2018progress}.
    \item \textbf{CMP}: the proposed local-learning architecture with movement-based plasticity regulation.
\end{itemize}

The following subsection reports the quantitative comparison across these three approaches using the continual-learning metrics introduced in Section~\ref{subsec:protocol}.
\subsection{Baseline comparison: naive fine-tuning and online EWC}
\label{subsec:ewc_baseline}

To evaluate whether the proposed plasticity-regulation mechanism provides improvements beyond standard continual-learning approaches, we compare CMP against a parameter-matched ($\sim$6.5M parameters) byte-level Transformer trained under the identical domain-incremental protocol described in Section~\ref{subsec:protocol}. Two Transformer baselines are considered: naive sequential fine-tuning without any forgetting mitigation and Online Elastic Weight Consolidation (Online EWC)~\citep{schwarz2018progress}, a scalable variant of the original EWC method proposed by Kirkpatrick et al.~\citep{kirkpatrick2017overcoming}.

All models are trained using the same domain sequence, optimization schedule, evaluation protocol, and data budget. Matching both model capacity and training conditions allows differences in continual-learning performance to be attributed primarily to the learning mechanisms rather than computational resources.

Before conducting the comparison, an initial baseline experiment was performed using the original 150,000-byte training budget. Under these conditions, the Transformer substantially overfit because the limited dataset required repeated presentation of the same samples, resulting in post-training BPB that exceeded the corresponding bigram baseline across all domains. Since the accuracy audit described in Section~\ref{subsec:audit} was not satisfied, these runs were excluded from the continual-learning comparison because the resulting BWT values would primarily reflect unsuccessful language modeling rather than catastrophic forgetting. Increasing the training budget to 3.8 million bytes resolved this issue, after which all Transformer baselines consistently passed the accuracy audit.

\begin{table}[t]
    \centering
    \caption{Continual-learning comparison using matched model capacity ($\sim$6.5M parameters), identical training protocol, and a shared 3.8M-byte training budget. Lower BWT indicates less forgetting.}
    \label{tab:baseline}
    \small
    \begin{tabular}{@{}lcc@{}}
        \toprule
        \textbf{Method} &
        \textbf{BWT} &
        \textbf{Relative forgetting} \\
        \midrule
        Naive fine-tuning (Transformer) &
        $+2.7897$ &
        Baseline \\
        Online EWC (Transformer) &
        $+2.2457$ &
        19.5\% reduction \\
        CMP (proposed) &
        $\mathbf{+0.1482}$ &
        \textbf{94.7\% reduction} \\
        \bottomrule
    \end{tabular}
\end{table}

\begin{figure}[t]
    \centering
    \includegraphics[width=0.60\linewidth]{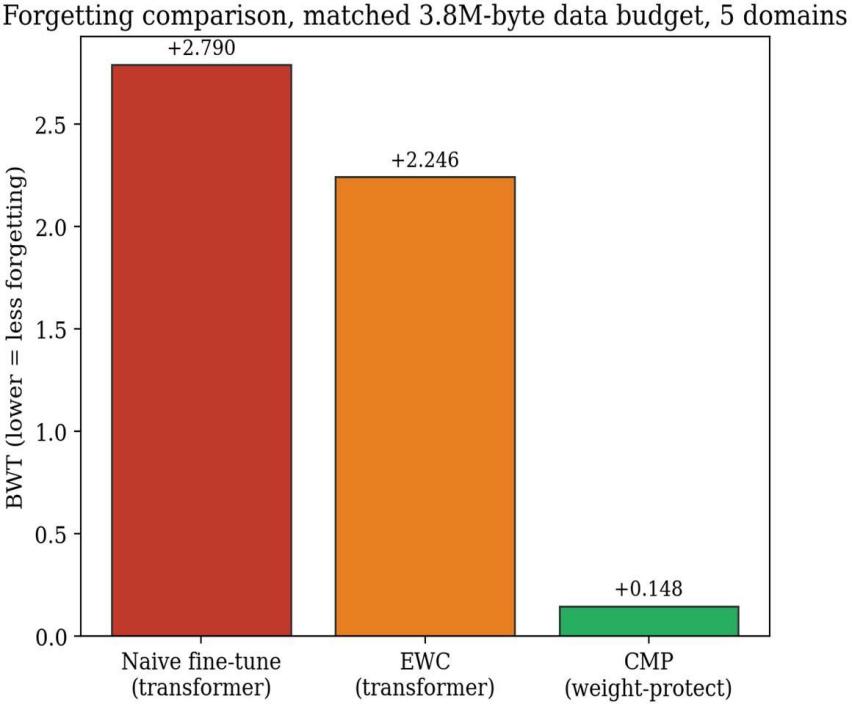}
    \caption{Backward Transfer (BWT) comparison under matched model capacity and a shared 3.8M-byte training budget. Lower values indicate reduced forgetting.}
    \label{fig:baseline}
\end{figure}

Table~\ref{tab:baseline} and Figure~\ref{fig:baseline} show that naive sequential fine-tuning exhibits substantial forgetting under the domain-incremental evaluation protocol. Online EWC reduces forgetting relative to naive fine-tuning, lowering BWT from $+2.7897$ to $+2.2457$, while CMP achieves the lowest BWT of $+0.1482$. Relative to naive fine-tuning, this corresponds to a 94.7\% reduction in forgetting, and a 93.4\% reduction relative to Online EWC under matched training conditions.

These results indicate that the proposed movement-based plasticity regulation substantially reduces interference between sequentially learned domains within the evaluation protocol considered in this study. While the comparison is limited to the selected benchmark and parameter-matched Transformer baselines, the observed improvement is consistent across the experimental conditions evaluated.

Additional validation on standardized continual-learning benchmarks, larger-scale language modeling tasks, and alternative architectures would further clarify the generality of the proposed approach.

To ensure a fair comparison, the CMP result reported in Table~\ref{tab:baseline} differs from the three-seed mean reported in Table~\ref{tab:replication}. The value in Table~\ref{tab:baseline} ($+0.1482$) was obtained after retraining CMP using the same 3.8M-byte data budget adopted for the Transformer baselines. This larger budget was introduced because the original 150,000-byte budget did not allow the Transformer baselines to satisfy the accuracy audit described in Section~\ref{subsec:audit}. Using the CMP result obtained under the matched experimental conditions therefore ensures that all methods are compared using identical model capacity, training protocol, and data budget.

Under these matched conditions, CMP achieves substantially lower forgetting than either Transformer baseline. Specifically, CMP reduces BWT by approximately 94.7\% relative to naive sequential fine-tuning and by approximately 93.4\% relative to Online EWC, corresponding to roughly 18.8$\times$ and 15.2$\times$ lower forgetting, respectively.

These comparisons constitute the primary quantitative result of this work. The following subsection examines whether this improvement remains consistent under an alternative ordering of the training domains, thereby assessing the robustness of the observed continual-learning behavior beyond a single domain sequence.

\subsection{Domain-order control}
\label{subsec:order}

The previous experiments evaluate CMP using a single ordering of the 15-domain continual-learning benchmark. Since catastrophic forgetting may depend on the sequence in which domains are encountered, we perform an additional control experiment to assess the sensitivity of the proposed method to domain ordering.

Evaluating multiple permutations of the complete 15-domain benchmark was computationally infeasible within the scope of this study. Consequently, this analysis is performed on a representative five-domain subset consisting of Wikipedia, CPython source code, Shakespeare, the King James Bible, and Reuters news. Three orderings are evaluated: the original canonical ordering, the reverse ordering, and a randomly shuffled ordering. This experiment is intended to estimate the sensitivity of CMP to domain order rather than exhaustively characterize all possible permutations.

\begin{table}[t]
    \centering
    \caption{Sensitivity of CMP to domain ordering on a representative five-domain subset using the original 150,000-byte training budget. Lower BWT indicates reduced forgetting.}
    \label{tab:order}
    \small
    \begin{tabular}{@{}llc@{}}
        \toprule
        \textbf{Ordering} &
        \textbf{Domain sequence} &
        \textbf{BWT} \\
        \midrule
        Canonical &
        Wiki $\rightarrow$ CPython $\rightarrow$ Shakespeare $\rightarrow$ KJV $\rightarrow$ Reuters &
        $+0.2353$ \\
        Reverse &
        Reuters $\rightarrow$ KJV $\rightarrow$ Shakespeare $\rightarrow$ CPython $\rightarrow$ Wiki &
        $+0.4442$ \\
        Shuffled &
        Shakespeare $\rightarrow$ Wiki $\rightarrow$ Reuters $\rightarrow$ CPython $\rightarrow$ KJV &
        $+0.3343$ \\
        \bottomrule
    \end{tabular}
\end{table}

Table~\ref{tab:order} demonstrates that continual-learning performance varies with the ordering of the training domains. Among the three sequences evaluated, the canonical ordering yields the lowest forgetting, whereas the reverse ordering produces the highest BWT. Across these experiments, BWT ranges from approximately $+0.24$ to $+0.44$.

Although domain ordering influences performance, every ordering evaluated substantially outperforms the Online EWC baseline reported in Section~\ref{subsec:ewc_baseline}.

This experiment represents only a partial assessment of order sensitivity. A five-domain benchmark admits $5! = 120$ possible orderings, whereas only three were evaluated in this study. Consequently, the reported range should be interpreted as an estimate of the observed variability rather than a bound on the worst-case continual-learning performance.

\begin{figure}[t]
    \centering
    \includegraphics[width=0.70\linewidth]{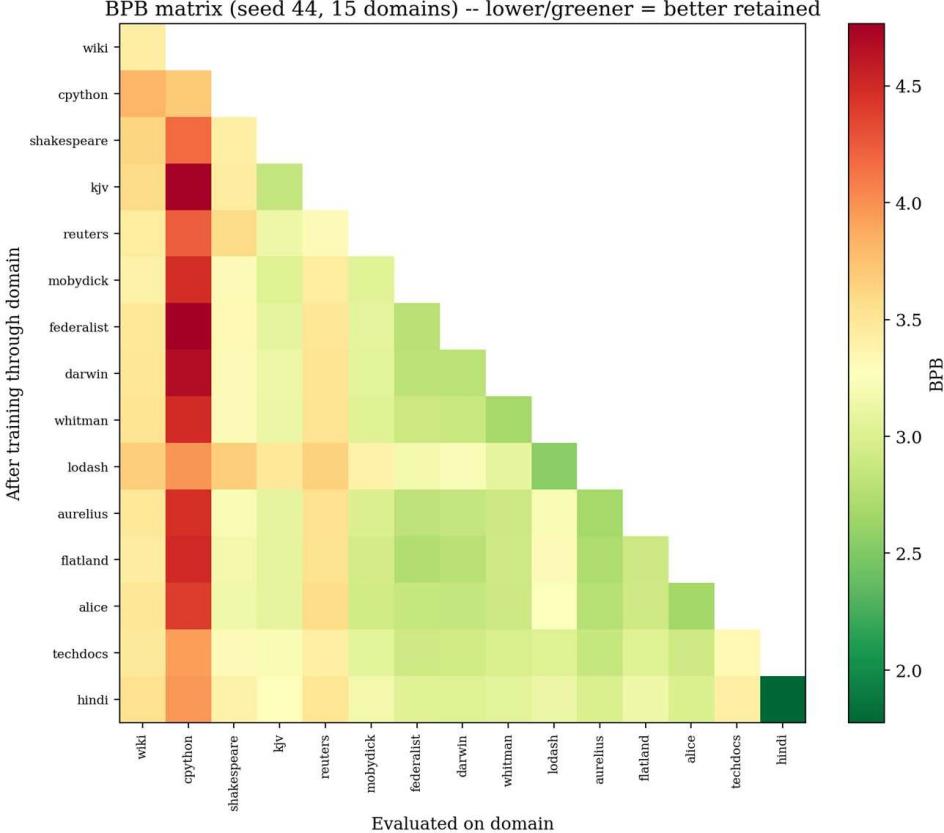}
    \caption{BPB evaluation matrix for the complete 15-domain benchmark (seed 44). Each row corresponds to the current training domain, and each column represents evaluation on a previously encountered domain. Lower BPB (green) indicates better retention, whereas higher BPB (red) indicates increased forgetting. The consistently elevated BPB values for the CPython domain suggest that it is retained less effectively than the remaining domains throughout the training sequence.}
    \label{fig:bpb-matrix}
\end{figure}

Figure~\ref{fig:bpb-matrix} provides a qualitative visualization of forgetting across the complete 15-domain benchmark. Most domains remain within a relatively narrow BPB range throughout continual training, whereas the CPython domain consistently exhibits higher BPB values, indicating greater susceptibility to forgetting under the evaluated protocol.

While all experiments presented thus far focus on byte-level language modeling, it remains unclear whether the observed continual-learning behavior is specific to text or reflects a more general property of the proposed learning mechanism. The following subsection therefore evaluates CMP on a standard continual-learning benchmark outside the language domain.

\subsection{Vision anchor: Split-MNIST}
\label{subsec:vision}

To investigate whether the proposed plasticity-regulation mechanism extends beyond language modeling, we evaluate CMP on the standard Split-MNIST continual-learning benchmark. Following the experimental setup adopted in this study, each MNIST image is serialized using a Hilbert-curve traversal and divided into the conventional five sequential tasks:
\[
0/1 \rightarrow 2/3 \rightarrow 4/5 \rightarrow 6/7 \rightarrow 8/9.
\]

The dense baseline and the sparse CMP configuration are evaluated across three independent random seeds (42, 43, and 44).

\begin{table}[t]
    \centering
    \caption{Split-MNIST continual-learning results using Hilbert-serialized inputs. Values report mean $\pm$ standard deviation across three random seeds.}
    \label{tab:vision}
    \small
    \begin{tabular}{@{}lcc@{}}
        \toprule
        \textbf{Condition} &
        \textbf{BWT} &
        \textbf{FWT} \\
        \midrule
        Dense (baseline) &
        $-0.0211 \pm 0.0004$ &
        $-0.0669 \pm 0.0013$ \\
        Sparse (CMP) &
        $-0.0207 \pm 0.0003$ &
        $-0.0666 \pm 0.0013$ \\
        \bottomrule
    \end{tabular}
\end{table}

Table~\ref{tab:vision} shows that the dense and sparse configurations produce nearly identical continual-learning performance on Split-MNIST. The differences in both BWT and FWT are substantially smaller than the observed variability across random seeds, and no consistent improvement attributable to the proposed plasticity-regulation mechanism is evident under this evaluation protocol.

Accordingly, we treat this experiment as a negative result rather than evidence supporting the proposed method. Although the current implementation does not demonstrate measurable improvements on Split-MNIST, reporting this outcome provides a more complete evaluation of the method than selectively presenting only favorable results.

The metrics reported in Table~\ref{tab:vision} are expressed in Bits-Per-Byte rather than classification accuracy because the experiments are conducted using the same autoregressive byte-prediction framework employed throughout this work. Consequently, these values should not be compared directly with the classification accuracies commonly reported for Split-MNIST continual-learning methods such as EWC, Synaptic Intelligence, or replay-based approaches.
\section{Discussion and Limitations}
\label{sec:discussion}

The results presented in this work demonstrate that CMP substantially reduces catastrophic forgetting under the evaluated continual-learning protocol while simultaneously highlighting several limitations that define the current scope of the proposed approach.

\paragraph{Predictive performance.}

Although CMP consistently reduces forgetting relative to the evaluated baselines, its single-domain language-modeling performance remains well below that of a comparably sized Transformer. Across the evaluated datasets, CMP achieves BPB values in the range of approximately 3.1--3.3, whereas modern Transformer architectures typically achieve substantially lower perplexity under comparable training conditions. Consequently, the principal contribution of this work concerns continual-learning behavior rather than state-of-the-art language-modeling accuracy. Bridging this performance gap while preserving the continual-learning properties demonstrated here remains an important direction for future research.

\paragraph{Sensitivity to domain ordering.}

The continual-learning performance of CMP depends on the order in which domains are presented. Across the three domain sequences evaluated in Section~\ref{subsec:order}, BWT ranges from approximately $+0.24$ to $+0.44$, with the canonical ordering producing the lowest forgetting. Because only three of the $5! = 120$ possible orderings of the five-domain subset were evaluated, these experiments should be interpreted as an estimate of order sensitivity rather than a comprehensive characterization of worst-case continual-learning performance.

\paragraph{Baseline scope.}

The principal continual-learning baseline employed in this work is Online Elastic Weight Consolidation (Online EWC)~\citep{schwarz2018progress}, which maintains a running estimate of parameter importance. This variant was selected because it scales naturally to long sequences of domains without requiring separate Fisher Information Matrices for every task, unlike the original formulation proposed by Kirkpatrick et al.~\citep{kirkpatrick2017overcoming}. The present study therefore evaluates CMP against Online EWC rather than the original EWC algorithm.

\paragraph{Evaluation benchmark.}

The 15-domain continual-learning benchmark introduced in this work was assembled specifically for this study and does not constitute a standardized benchmark. While the continual-learning metrics (Backward Transfer and Forward Transfer) are well established in the literature, the underlying collection of evaluation domains is custom. Additional validation on widely adopted continual-learning benchmarks would strengthen the generality of the reported findings.

\paragraph{Generalization beyond language modeling.}

The Split-MNIST experiments did not demonstrate a measurable advantage of the proposed plasticity-regulation mechanism over the dense baseline. Across three independent random seeds, the observed differences in BWT and FWT remained within the variability of the experiments, and no consistent improvement was observed. Accordingly, these results should be interpreted as a negative result rather than evidence that the proposed mechanism generalizes beyond language modeling.

\paragraph{Methodological contribution.}

The individual architectural components employed by CMP, including sparse relational representations, associative memory, and local delta-rule learning, build upon established ideas discussed in Section~\ref{sec:related}. The primary methodological contribution of this work is the integration of these components with a movement-based plasticity-regulation mechanism that estimates parameter importance from accumulated weight movement rather than gradient-derived quantities, together with an empirical evaluation of this approach under continual language-modeling conditions.

\paragraph{Future directions.}

The experiments reported in Appendix~C indicate that alternative architectural variants can achieve substantially lower single-domain BPB than the configuration presented in the main paper. However, combining these improvements with the proposed continual-learning architecture did not improve overall performance under the evaluated setting (Appendix~D). Understanding how stronger predictive representations can be integrated with movement-based plasticity regulation without sacrificing continual-learning behavior remains an open research question.

Overall, the results suggest that movement-based plasticity regulation is a promising direction for continual learning within local-learning architectures. At the same time, broader evaluation on standardized benchmarks, stronger language-modeling baselines, additional continual-learning methods, and larger-scale models will be necessary to determine the generality and practical applicability of the proposed approach.
\section{Conclusion}
\label{sec:conclusion}

This work investigated whether a local-learning architecture with sparse representations and movement-based plasticity regulation can mitigate catastrophic forgetting in continual language modeling without relying on end-to-end backpropagation or replay. To evaluate this hypothesis, we introduced CMP and assessed its performance under a domain-incremental continual-learning protocol using established transfer metrics together with comparisons against parameter-matched Transformer baselines.

Across the evaluated experiments, CMP consistently exhibited substantially lower forgetting than both naive sequential fine-tuning and Online Elastic Weight Consolidation (Online EWC). The observed reduction in Backward Transfer was reproduced across multiple random seeds and remained evident under alternative domain orderings, suggesting that the proposed plasticity-regulation mechanism contributes meaningfully to continual learning within the evaluated setting.

At the same time, the present study identifies several important limitations. CMP does not yet achieve language-modeling accuracy comparable to modern Transformer architectures, the evaluation is conducted on a custom continual-learning benchmark, and the proposed method does not demonstrate measurable improvements on Split-MNIST. These results indicate that reduced forgetting alone is insufficient to establish a broadly competitive learning architecture.

Future work should investigate how movement-based plasticity regulation can be combined with stronger predictive architectures while preserving the continual-learning behavior demonstrated in this study. Broader evaluation on standardized continual-learning benchmarks, larger-scale language models, and additional baseline methods will also be necessary to assess the generality of the proposed approach.

\appendix

\section{Hyperparameters}
\label{app:hparams}

Table~\ref{tab:hparams} summarizes the hyperparameters used for the flagship CMP configuration reported throughout the main paper. Unless otherwise stated, all experiments employ these settings.

\begin{table}[t]
    \centering
    \caption{Hyperparameters used for the flagship CMP configuration.}
    \label{tab:hparams}
    \small
    \begin{tabular}{@{}ll@{}}
        \toprule
        \textbf{Parameter} & \textbf{Value} \\
        \midrule
        Representation dimension ($r$) & 1024 \\
        Batch size & 64 \\
        Sequence length & 256 \\
        Training steps per domain & 6000 \\
        Data budget & 150,000 bytes (Sections~4.1--4.5) \\
        & 3,800,000 bytes (Transformer baseline comparison) \\
        Learning rate ($\eta$) & 0.12 \\
        Plasticity-regulation coefficient ($\lambda$) & 5.0 \\
        Sparsity ($k$) & 64 \\
        Buffer size & 64 \\
        Register size & 256 \\
        \bottomrule
    \end{tabular}
\end{table}

The sparsity level, buffer capacity, and register capacity scale proportionally with the representation dimension. The values shown correspond to the flagship configuration with $r=1024$.

\section{Per-domain accuracy audit}
\label{app:audit}

Section~\ref{subsec:audit} establishes that continual-learning metrics are only meaningful when the underlying language model learns beyond a trivial baseline. Accordingly, we compare the post-training BPB of CMP against a bigram frequency model for every evaluation domain and every random seed.

Across all fifteen domains and all three random seeds (42, 43, and 44), CMP consistently achieves lower BPB than the corresponding bigram baseline. Improvements range from approximately 0.41 to 1.22 BPB depending on the evaluation domain.

The complete per-domain evaluation tables are included in the released experimental results accompanying this work.

\section{Historical depth-line result}
\label{app:depth}

Prior to the development of the architecture presented in the main paper, we investigated a separate family of locally trained architectures based on stacked multiplicative binding layers. Unlike CMP, these models did not include competitive memory, predictive coding, or movement-based plasticity regulation, and were evaluated only on single-domain language modeling.

The strongest configuration in this earlier architecture family achieved approximately 2.49--2.51 BPB on single-domain byte prediction using four stacked binding blocks, with results replicated across three random seeds. A control experiment that randomized the deepest binding layer confirmed that the observed improvement resulted from the additional representational depth rather than increased parameter count alone.

These experiments motivated subsequent attempts to combine deeper local representations with the continual-learning mechanisms proposed in this work. The outcome of that integration attempt is reported in Appendix~\ref{app:merge}.
\section{Depth--Hierarchy Integration Attempt}
\label{app:merge}

To investigate whether stronger single-domain predictive performance could be combined with the continual-learning architecture proposed in this work, we integrated the depth-only architecture described in Appendix~\ref{app:depth} with the CMP hierarchy and competitive memory system. Specifically, the original single relational-binding stage was replaced with a four-block stacked binding architecture while leaving the hierarchy, memory, predictive-coding, and local-learning components unchanged.

Under matched training conditions, the integrated architecture achieved a BPB of 3.48 compared with 3.27 for the flagship CMP configuration, indicating that the modification did not improve language-modeling performance.

One plausible explanation is that the depth-only architecture produces dense intermediate representations through conventional normalization, whereas the competitive memory and downstream processing stages in CMP assume sparse representations generated by the kWTA operator. The resulting mismatch may reduce the quality of associative-memory retrieval and subsequent hierarchical processing. A second possibility is that introducing additional representational stages increases competition among multiple readout pathways that ultimately share the same local prediction error.

These explanations are hypotheses derived from the observed behavior rather than experimentally verified causes. Accordingly, we report this experiment as a negative result that identifies a potential direction for future investigation rather than evidence against deeper local-learning architectures.

\section{Preliminary scaling experiment}
\label{app:scale}

We performed a preliminary investigation of model scaling by increasing the representation dimension from 1024 to 2600, corresponding to an increase in model size from approximately 6.4 million to 29.8 million parameters.

Under the original 150,000-byte training budget, this five-fold increase in parameter count produced only negligible improvement. Subsequent analysis indicated that the limited data budget required extensive repetition of the training data, substantially reducing the amount of novel information available to the larger model.

Repeating the comparison using the corrected 3.8M-byte training budget resulted in a modest improvement, with BWT decreasing from $+0.1482$ for the 6.4M-parameter model to $+0.1397$ for the 29.8M-parameter model. While this result suggests that increased model capacity becomes beneficial once the data bottleneck is alleviated, the observed improvement remains considerably smaller than the effect of increasing the training budget itself.

This experiment represents only a preliminary exploration of scaling behavior. A systematic scaling-law analysis across model size, data budget, and training compute remains an important direction for future work.

\section{Code and Data Availability}
\label{app:availability}

The continual-learning text corpus used in this work was assembled from publicly available public-domain and open-source sources, as described in Section~\ref{subsec:protocol}. The vision experiments use the standard Split-MNIST benchmark constructed from the MNIST dataset.

The source code and experimental framework are not publicly available at the time of publication. They are available from the authors upon reasonable request for research purposes.

\section{Acknowledgments}
\label{app:acknowledgments}

The authors acknowledge the Google for Startups Cloud Program for providing cloud-computing credits that supported the experiments reported in this work. Experiments were conducted using NVIDIA A100, NVIDIA T4, and NVIDIA L4 GPUs.

The authors also thank Arjit Saxena (Resolute Labs) for mentorship and technical guidance during the development of this research.

\bibliography{references}

\end{document}